\pdfoutput=1

\documentclass[11pt]{article}

\usepackage[]{emnlp2022}

\usepackage{times}
\usepackage{latexsym}
\usepackage{graphicx}
\usepackage{amsmath}
\usepackage{booktabs}
\usepackage{tabularx}

\usepackage[T1]{fontenc}

\usepackage[utf8]{inputenc}

\usepackage{microtype}

\definecolor{Orange}{RGB}{255,140,0}
 
\newcommand{\figref}[1]{Figure~\ref{#1}}
\newcommand{\tabref}[1]{Table~\ref{#1}}
\newcommand{\secref}[1]{Section~\ref{#1}}

\usepackage{inconsolata}
\usepackage{titlesec}
\titlespacing*{\paragraph}{\parindent}{0ex}{1ex}
\usepackage{multirow}

\hypersetup{
    pdftitle={Context Matters for Image Descriptions for Accessibility: Challenges for Referenceless Evaluation Metrics},    
    pdfauthor={Elisa Kreiss, Cynthia Bennett, Shayan Hooshmand, Eric Zelikman, Meredith Ringel Morris, Christopher Potts},
    pdfkeywords={natural language processing, image accessibility, computer vision},
    pdfstartpage={1},
    pdflang={en}
}

\title{Context Matters for Image Descriptions for Accessibility:\\Challenges for Referenceless Evaluation Metrics}

\author{
  {\bf Elisa Kreiss}\thanks{\ \ Corresponding author: \url{ekreiss@stanford.edu}} \\
  Stanford University \\
  \And
  {\bf Cynthia Bennett} \\
  Google Research \\
  \And
  {\bf Shayan Hooshmand} \\
  Columbia University \\
  \AND
  {\bf Eric Zelikman} \\
  Stanford University \\
  \And
  {\bf Meredith Ringel Morris} \\
  Google Research \\
  \And
  {\bf Christopher Potts} \\
  Stanford University \\
  }

\begin{document}
\maketitle
\begin{abstract}
Few images on the Web receive alt-text descriptions that would make them accessible to blind and low vision (BLV) users. Image-based NLG systems have progressed to the point where they can begin to address this persistent societal problem, but these systems will not be fully successful unless we evaluate them on metrics that guide their development correctly. Here, we argue against current \emph{referenceless} metrics -- those that don't rely on human-generated ground-truth descriptions -- on the grounds that they do not align with the needs of BLV users. The fundamental shortcoming of these metrics is that they do not take context into account, whereas contextual information is highly valued by BLV users. To substantiate these claims, we present a study with BLV participants who rated descriptions along a variety of dimensions. An in-depth analysis reveals that the lack of context-awareness makes current referenceless metrics inadequate for advancing image accessibility. As a proof-of-concept, we provide a contextual version of the referenceless metric CLIPScore which begins to address the disconnect to the BLV data. 
An accessible HTML version of this paper is available at \url{https://elisakreiss.github.io/contextual-description-evaluation/paper/reflessmetrics.html}
\end{abstract}

\section{Introduction}

In the pursuit of ever more powerful image description systems, we need evaluation metrics that provide a clear window into model capabilities. At present, we are seeing a rise in \emph{referenceless} (or reference-free) metrics \cite{hessel2021clipscore,lee2021qace,lee2021umic,feinglass2021smurf}, building on prior work in domains such as machine translation \cite{lo2019yisi, zhao2020limitations} and summarization \cite{louis2013automatically, peyrard2018objective,gao-etal-2020-supert,deutsch2021towards}. These metrics seek to estimate the quality of a text corresponding to an image without requiring ground-truth labels (i.e., reference descriptions), or crowd worker judgments. 
Thus, they offer the promise of quick and efficient evaluation of image description models, and are even suggested to be more reliable than existing reference-based metrics \cite{kasai2021bidimensional,kasai2021transparent}. Here, we investigate the value of such metrics for a high social-impact domain: assessing the usefulness of image descriptions for blind and low vision (BLV) users.

\begin{figure}[t!]
  \centering
  \includegraphics[width=0.45\textwidth]{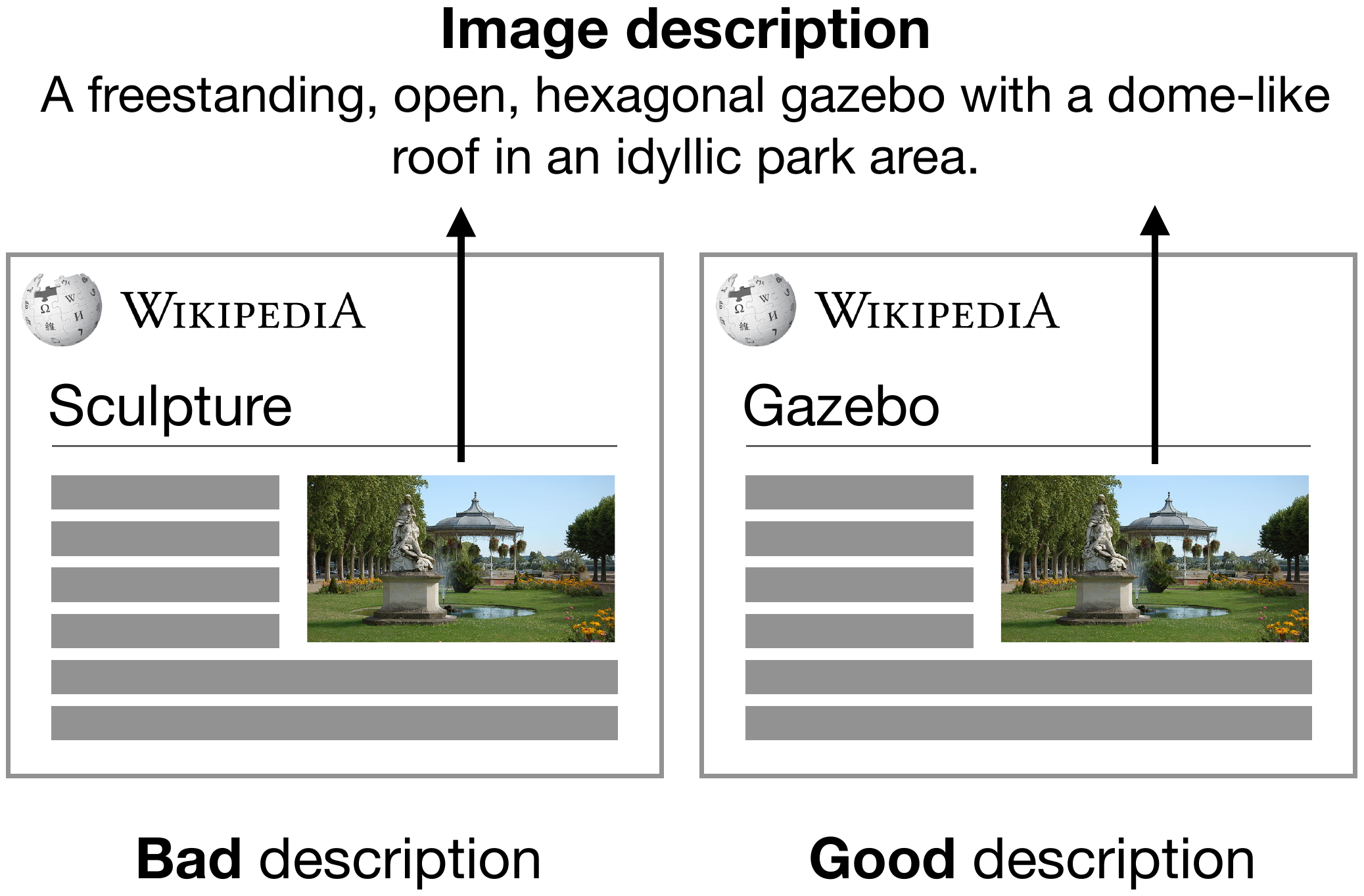}
  \caption{Whether an image description makes an image accessible depends on the context in which the image appears. Referenceless metrics like CLIPScore can't capture such context-sensitivity. We provide experimental evidence with blind and low vision (BLV) participants that this makes current referenceless metrics insufficient for evaluating image description quality.} 
  \label{fig:contextual-relevance}
\end{figure}

Automatically generating descriptions to make images accessible is an important goal: though images are omnipresent in digital communication \cite{hackett2003accessibility, bigham2006webinsight, buzzi2011web, voykinska2016how, gleason2019it}, user-generated descriptions are rare \cite{gleason2020twitter}, which has serious implications for BLV users \cite{morris2016most}. Can referenceless metrics help guide models to generate descriptions that align with what BLV users value? 

Studies with BLV users emphasize the importance of the \emph{context} in which an image appears. For example, while people's clothing is highly relevant when browsing shopping websites, their identities become central when reading the news \cite{stangl2021going, muehlbradt2022what,stangl2020person}. 
Not only the domain but also the immediate context matters for selecting what is relevant. Consider the image in \figref{fig:contextual-relevance}, showing a park with a gazebo in the center and a sculpture on a pedestal in the foreground. A description written for the image's occurrence in the Wikipedia article of gazebos (``A freestanding, open, hexagonal gazebo with a dome-like roof in an idyllic park area.'')\ is unhelpful if the image instead appears in the article on sculpture. This simple example illustrates that context could play a central role in the assessment of description quality. 

In this work, we report on studies with sighted and BLV participants that seek to provide rich, multidimensional information about what people value in image descriptions for accessibility. In contrast to current practices, we elicit and evaluate image descriptions within contexts the images could appear in, here Wikipedia articles. We find that, for both sighted and BLV participants, the description's relevance to the context is a major driver of their overall assessments. 

We then use this experimental data to evaluate two very different referenceless metrics: CLIPScore \cite{hessel2021clipscore}, which assesses a description's quality relative to its associated image, and SPURTS \citep{feinglass2021smurf}, which relies only on linguistic properties of the text. By their very design, these metrics don't capture the effects of context seen in our user studies, since they treat description evaluation as a context-less problem. This shortcoming goes undetected on most existing datasets and previously conducted human evaluations, which presume that image descriptions are context-independent, but it is immediately apparent in our evaluations.

These results suggest that current referenceless metrics are not reliable guides due to their lack of context integration, but offers a path forward: perhaps referenceless metrics can be modified to include this missing context. As a proof-of-concept, we show that a context-sensitive adaptation of CLIPScore results in improved correlations with human judgments -- a promising signal for the development of future context-sensitive referenceless metrics.

\section{Background}

\begin{figure*}[t!]
  \centering
  \includegraphics[width=0.95\textwidth]{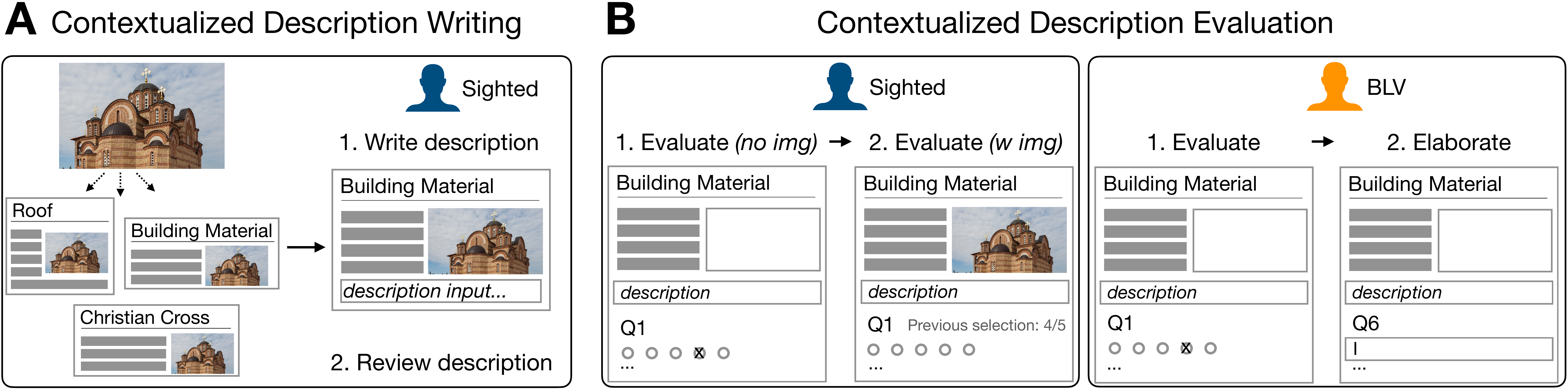}
  \caption{Experimental design overview consisting of two main phases: (A) eliciting descriptions written for images occurring within varying contexts, (B) obtaining detailed evaluations of those descriptions from sighted and BLV participants. These evaluations give insights into the role that context needs to play for providing useful descriptions, and function as the gold standard that the results from referenceless metrics are then compared to.} 
  \label{fig:expdesign}
\end{figure*}

\subsection{Image Accessibility}
 
Screen readers provide auditory and braille access to Web content. To make images accessible in this way, screen readers use image descriptions embedded in HTML \texttt{alt} tags. However, such descriptions are rare. While frequently visited websites are estimated to have about 72\% coverage \cite{guinness2018caption}, this drops to less than 6\% on English-language Wikipedia \cite{kreiss2022concadia} and to 0.1\% on English-language Twitter \cite{gleason2019it}. This has severe implications especially for BLV users who have to rely on such descriptions to engage socially \cite{morris2016most, macleod2017understanding, buzzi2011web, voykinska2016how} and stay informed \cite{gleason2019it, morris2016most}.

Moreover, these coverage estimates are based on any description being available, without regard for whether the descriptions are useful. Precisely what constitutes a useful description is still an underexplored question. A central finding from work with BLV users is that one-size-fits-all image descriptions don't address image accessibility needs \cite{stangl2021going, muehlbradt2022what, stangl2020person}. \citet{stangl2021going} specifically tested the importance of the \emph{scenario} -- the source of the image and the informational goal of the user -- by placing each image within different source domains (e.g., news or shopping website) which were associated with specific goals (e.g., learning or browsing for a gift). They find that BLV users have certain description preferences that are stable across scenarios (e.g., people's identity and facial expressions, or the type of location depicted), whereas others are scenario-dependent (e.g., hair color). 
We extend this previous work by keeping the scenario stable but varying the immediate context the image is embedded in.

Current referenceless metrics take the one-size-fits-all approach. We explicitly test whether this is sufficient to capture the ratings provided by BLV users when they have access to the broader context.

\subsection{Image-based Text Evaluation Metrics}

There are two evaluation strategies for automatically assessing the quality of a model's generated text from images: \emph{reference-based} and \emph{referenceless} (or reference-free) metrics.

Reference-based metrics rely on human-created ground-truth texts associated with each image.
The candidate text generated by the model is then compared with those ground-truth references, returning a similarity score. A wide variety of scoring techniques have been explored. Examples are BLEU \cite{papineni2002bleu}, CIDEr \cite{vedantam2015cider}, SPICE \cite{anderson2016spice}, ROUGE \cite{lin2004rouge}, and BERTscore \cite{zhang2019bertscore}. The more references are provided, the more reliable the scores, which requires datasets with multiple high-quality annotations for each image. Such datasets are expensive and difficult to obtain.

As discussed above, referenceless metrics dispense with the need for ground-truth reference texts. Instead, text quality is assessed based either on how the text relates to the image content \cite{hessel2021clipscore,lee2021umic,lee2021qace} or on text quality alone \cite{feinglass2021smurf}. As a result, these metrics can in principle be used anywhere without the need for an expensive annotation effort. 

How the score is computed varies between metrics. CLIPScore \cite{hessel2021clipscore} and UMIC \cite{lee2021umic} pose a classification problem where models are trained contrastively on compatible and incompatible image--text pairs. A higher score for a given image and text as input then corresponds to a high compatibility between them. QACE provides a high score if descriptions and images give similar answers to the same questions \cite{lee2021qace}. SPURTS is a referenceless metric which judges text quality solely based on text-internal properties that can be conceptualized as maximizing unexpected content \cite{feinglass2021smurf}. SPURTS was originally proposed as part of the metric SMURF,  which additionally contains a reference-based component, specifically designed to capture the semantics of the description. However, \citeauthor{feinglass2021smurf} find that SPURTS alone already seems to approximate human judgments well, which makes it a relevant referenceless metric to consider. While varying in their approach, all current referenceless metrics share that they treat image-based text generation as a context-independent problem.

Reference-based metrics have the \emph{potential} to reflect context-dependence, assuming the reference texts are created in ways that engage with the image's context. Referenceless methods are much more limited in this regard: if a single image--description pair should receive different scores in different contexts, but the metric operates only on image--description pairs, then the metric will be intrinsically unable to provide the desired scores.

\section{Experiment: The Effect of Context on Human Image Description Evaluation}

Efforts to obtain and evaluate image descriptions through crowdsourcing are mainly conducted out-of-context: images that might have originally been part of a tweet or news article are presented in isolation to obtain a description or evaluation thereof. Following recent insights on the importance of the domains an image appeared in \cite{stangl2021going, muehlbradt2022what, stangl2020person}, we seek to understand the role of context in shaping how people evaluate descriptions. \figref{fig:expdesign} provides an overview of the two main phases. Firstly, we obtained contextual descriptions by explicitly varying the context each image could occur in (\figref{fig:expdesign}A). We then explored how context affects sighted and BLV users' assessments of descriptions along a number of dimensions (\figref{fig:expdesign}B). Finally, in \secref{sec:metric-assess}, we compare these contextual evaluations with the results from the referenceless metrics CLIPScore \cite{hessel2021clipscore} and SPURTS \cite{feinglass2021smurf}.

\subsection{Data}

To investigate the effect of context on image descriptions, we designed a dataset where each image was paired with three distinct contexts, here Wikipedia articles. For instance, an image of a church was paired with the first paragraphs of the Wikipedia articles on \emph{Building material}, \emph{Roof}, and \emph{Christian cross}. Similarly, each article appeared with three distinct images. The images were made publicly available through Wikimedia Commons. Overall, we obtained 54 unique image--context pairs, consisting of 18 unique images and 17 unique articles. The dataset, experiments used for data collection, and analyses are 
made available.\footnote{\url{https://github.com/elisakreiss/contextual-description-evaluation}}

\subsection{Contextual Description Writing}

\begin{figure*}[t!]
  \centering
  \includegraphics[width=0.95\textwidth]{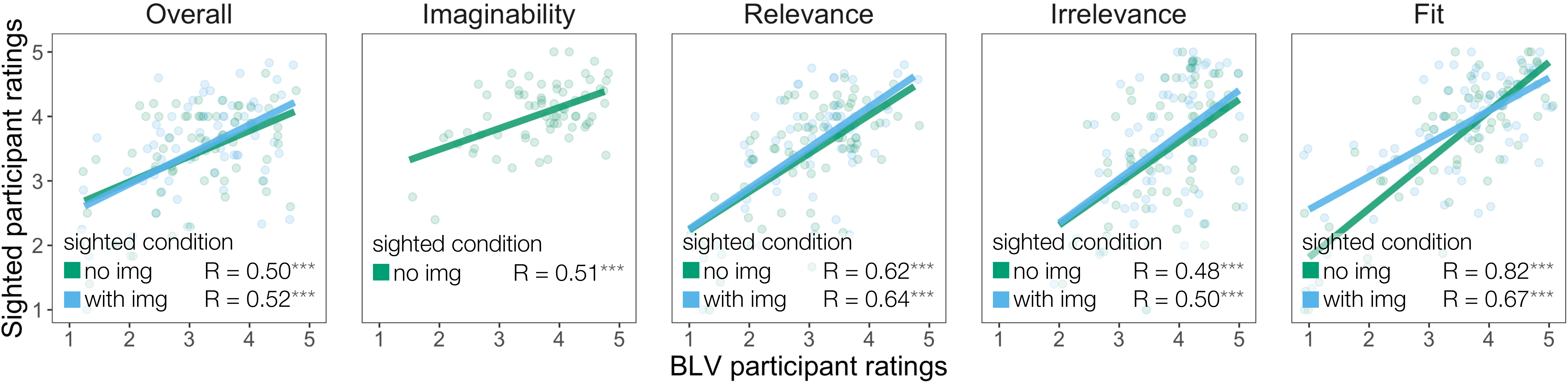}
  \caption{Correlation of BLV and sighted participant ratings across questions. Sighted participants provided ratings twice -- before seeing the image (in green) and after (in blue). Each point denotes the average rating for a description. The Pearson correlations (R) are all statistically significant, as indicated by the asterisks. For all questions, higher ratings are associated with higher quality descriptions.} 
  \label{fig:corr-blv-sighted}
\end{figure*}
 
In our first experiment, participants sought to write descriptions that could make images accessible to users who can't see them. 

\paragraph{Task} Each participant went through a brief introduction explaining the challenge and purpose of image descriptions and was then shown six distinct articles, each of them containing a different image they were asked to describe. To enable participants to judge their descriptions, the description then replaced the image and participants could choose to edit their response before continuing. The task did not contain any guidance on which information should or should not be included in the description. Consequently, any context-dependence is simply induced by presenting the images within contexts (Wikipedia articles) instead of in isolation.

\paragraph{Participants and Exclusions} We recruited 74 participants on Amazon's Mechanical Turk. We excluded six participants who indicated confusion about the task in the post-questionnaire and one for whom the experiment didn't display properly. Overall, each image--article pair received on average five descriptions.

\paragraph{Results} After exclusions, we obtained 272 descriptions that varied in length between 13 and 541 characters, with an average of 24.9 words. We evaluate to what extent the description content was affected by the image context based on the following human subject evaluation experiment.

\subsection{Contextual Description Evaluation}

After obtaining contextual image descriptions, we designed a description evaluation study which we conducted with BLV as well as sighted participants. Both groups can provide important insights. We consider the ratings of BLV participants as the primary window into accessibility needs. However, sighted participant judgments can complement these results, in particular by helping us determine whether a description is true for an image. Furthermore, the sighted participants' intuitions about what makes a good description are potentially informative since sighted users are usually the ones providing image descriptions. 

\paragraph{Task} Sighted as well as BLV participants rated each image description as it occurred within the respective Wikipedia article. To get a better understanding of the kinds of content that might affect description quality, each description was evaluated according to five dimensions: \emph{Overall} quality, \emph{Imaginability} of the image from the description, \emph{Relevance} and \emph{Irrelevance} of the mentioned details, and image \emph{Fit} to the article.

The \emph{Imaginability} and \emph{(Ir)Relevance} questions are designed to capture two central aspects of description content. While \emph{Imaginability} has no direct contextual component, \emph{Relevance} and \emph{Irrelevance} specifically ask about the contextually determined aspects of the description. These dimensions give us insights into the importance of context in the \emph{Overall} description quality ratings.

Responses were provided on 5-point Likert scales. In addition to 17 critical trials each participant completed, we further included two trials with descriptions carefully constructed to exhibit for instance low vs.~high context sensitivity. These trials allowed us to ensure that the questions and scales were interpreted as intended by the participants. Overall, each participant completed 19 trials, where each trial consisted of a different article and image. Trial order and question order were randomized between participants to avoid potential ordering biases.

\subsubsection{Sighted Participants}\label{sec:exp-eval-sighted}

\paragraph{Task} To ensure high data quality, sighted participants were asked a reading comprehension question before starting the experiment, which also familiarized them with the overall topic of image accessibility. If they passed, they could choose to enter the main study, otherwise they exited the study and were only compensated for completing the comprehension task. 

In each trial, participants first saw the Wikipedia article, followed by an image description. This \emph{no image} condition can be conceptualized as providing sighted participants with the same information as a BLV user. 
They then responded to the five questions and were asked to indicate if the description contained false statements or discriminatory language. After submitting the response, the image was revealed and participants responded again to four of the five questions. The \emph{Imaginability} question was omitted since it isn't clearly interpretable once the image is visible. Their previous rating for each question was made available to them so that they could reason about whether they wanted to keep or change their response.

\paragraph{Participants and Exclusions} 79 participants were recruited over Amazon's Mechanical Turk, 68 of whom continued past the reading comprehension question. We excluded eight participants since they spent less than 19 minutes on the task, and one participant whose logged data was incomplete. This resulted in 59 submissions for further analysis. 

\subsubsection{BLV Participants}

The 68 most-rated descriptions across the 17 Wikipedia articles and 18 images were then selected to be further evaluated by BLV participants.

\paragraph{Task} To provide BLV participants with the same information as sighted participants, they similarly started with the reading comprehension question before continuing to the main trials. After reading the Wikipedia article and the image description, participants first responded to the five evaluation dimensions. Afterwards, they provided answers to five open-ended questions about the description content. The main focus of the analysis presented here is on the Likert scale responses, but the open-ended explanations allow more detailed insights into description preferences. Each description was rated by exactly four participants.

\paragraph{Participants} 16 participants were recruited via 
email lists for BLV users, and participants were unknowing about the purpose of the study. Participants self-described their level of vision as totally blind (7), nearly blind (3), light perception only (5), and low vision (1). 15 participants reported relying on screen readers (almost) always when browsing the Web, and one reported using them often. 

We enrolled fewer blind participants than sighted participants, as they are a low-incidence population, requiring targeted and time-consuming recruitment. For example, crowd platforms that enable large sample recruitment are inaccessible to blind crowd workers \citep{Vashistha-etal:2018}.

\subsubsection{Evaluation Results}

The following analyses are based on the 68 descriptions, comprising 18 images and 17 Wikipedia articles. Each description is evaluated according to multiple dimensions by sighted as well as BLV participants for how well the description serves an accessibility goal. 

\figref{fig:corr-blv-sighted} shows the correlation of BLV and sighted participant ratings across questions. We find that the judgments of the two groups are significantly correlated for all questions. The correlation is encouraging since it shows an alignment between the BLV participants' reported preferences and the sighted participants' intuitions. Whether sighted participants could see the image when responding didn't make a qualitative difference. The results further show that the dataset provides very poor to very good descriptions, covering the whole range of possible responses. This range is important for insights into whether a proposed evaluation metric can detect what makes a description useful.

We conducted a mixed effects linear regression analysis of the BLV participant judgments to investigate which responses are significant predictors of the overall ratings. We used normalized and centered fixed effects of the three content questions (\emph{Imaginability}, \emph{Relevance} and \emph{Irrelevance}), and random by-participant and by-description intercepts. If context doesn't affect the quality of a description, \emph{Imaginability} should be a sufficient predictor of the overall ratings. However, in addition to an effect of \emph{Imaginability} ($\beta = .42$, $\text{SE} = .06$, $p<.001$), we find a significant effect of \emph{Relevance} as well ($\beta = .44$, $\text{SE} = .05$, $p<.001$), suggesting that context plays an essential role in guiding what makes a description useful. This finding replicates with the sighted participant judgments.

\begin{figure}[t!]
  \includegraphics[width=0.95\linewidth]{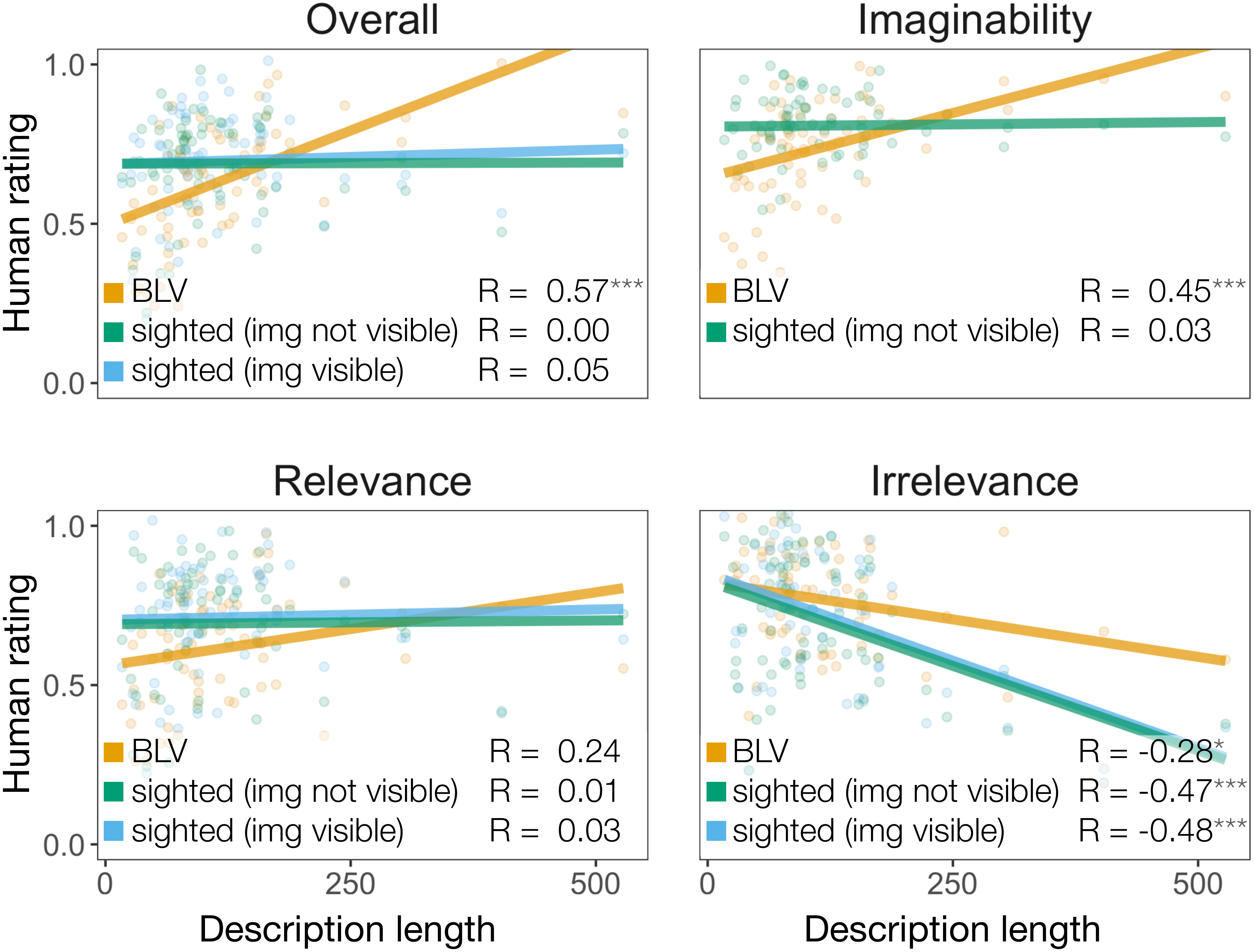}
  \caption{Correlation of BLV and sighted participant judgments with description length (in characters). Human judgments are rescaled to the zero to one range.} 
  \label{fig:length-corr}
\end{figure}

A case where BLV and sighted participant ratings diverge is in the effect of description length (\figref{fig:length-corr}). While longer descriptions tend to be judged overall more highly by BLV participants, there is no such correlation for sighted participants. This finding contrasts with popular image description guidelines, which often advocate for shorter descriptions.\footnote{E.g., \url{https://webaim.org/techniques/alttext/}} The lack of correlation between sighted participant ratings and description length might be linked to this potential misconception.

\section{Referenceless Metrics for Image Accessibility}\label{sec:metric-assess}

Referenceless metrics have been shown to correlate well with how sighted participants judge description quality when descriptions are written and presented out-of-context \cite{hessel2021clipscore,feinglass2021smurf,lee2021umic,kasai2021transparent}. While image accessibility is one of the main goals referenceless metrics are intended to facilitate \cite{kasai2021bidimensional,hessel2021clipscore,kasai2021transparent}, it remains unclear whether they can approximate the usefulness of a description for BLV users. Inspired by recent insights into what makes a description useful, we argue that the inherently decontextualized nature of current referenceless metrics makes them inadequate as a measure of image accessibility. We focus on two referenceless metrics to support these claims: CLIPScore \cite{hessel2021clipscore} and SPURTS \cite{feinglass2021smurf}. Appendix~\ref{app:others} briefly considers other referenceless metrics.

\begin{figure}[t!]
  \includegraphics[width=0.48\textwidth]{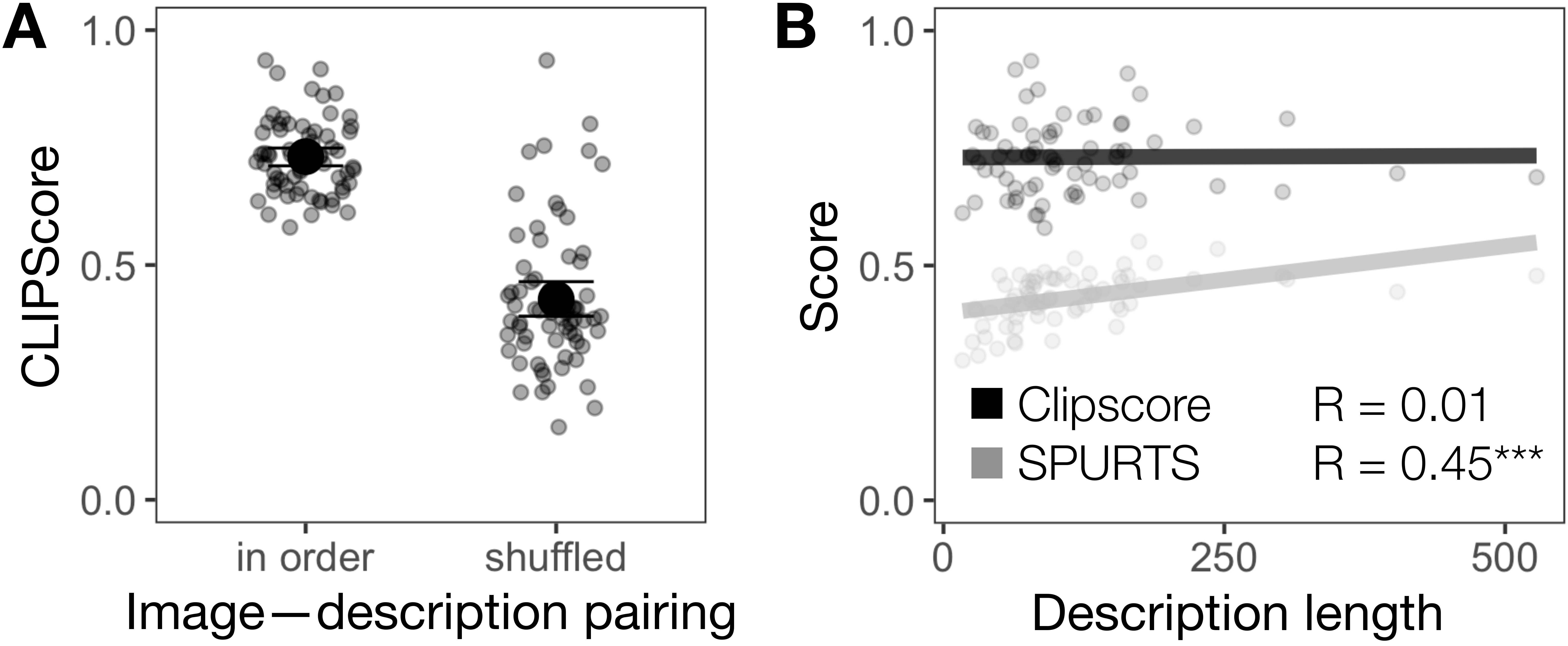}
  \caption{Analyses of the capabilities of referenceless metrics. (A) CLIPScore can pick out whether a written description is compatible with the image. When shuffling image--description pairs, the average CLIPScore drops from 0.73 to 0.43. SPURTS can't make this distinction due to its image-independence. (B) Longer descriptions are associated with higher scores of SPURTS but not CLIPScore.} 
  \label{fig:clipscore-truthfulness}
\end{figure}

\begin{figure*}[t!]
  \centering
  \includegraphics[width=0.95\textwidth]{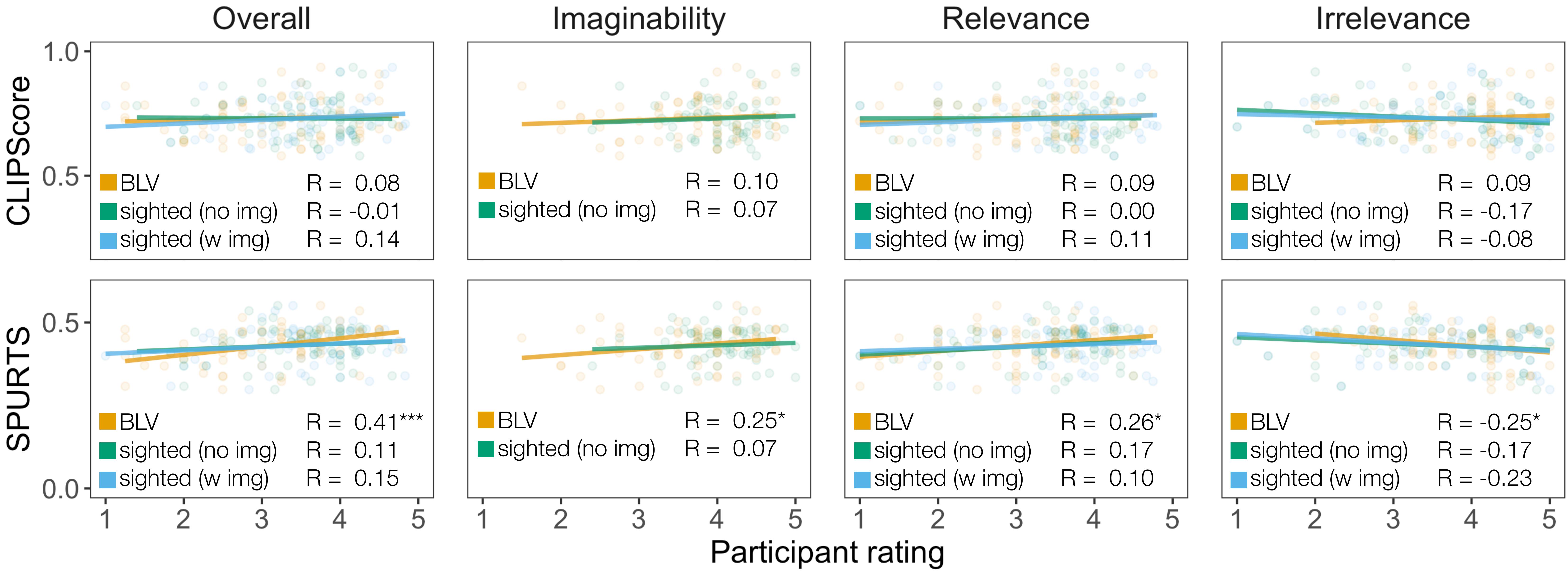}
  \caption{Correlations of CLIPScore (top row) and SPURTS (bottom row) with ratings 
  of
  sighted and BLV participants. The Pearson correlations are computed over the human evaluators' average per-description rating. Sighted participants responded to the questions twice; once without (in green) and after seeing the image (in blue). There are no significant correlations between CLIPScore and human ratings. SPURTS correlates significantly with all responses provided by BLV participants but negatively with \emph{Irrelevance} and not at all with sighted participant ratings, indicating a fundamental mismatch.} 
  \label{fig:clipscore-blvsighted}
\end{figure*}

CLIPScore uses the similarity of CLIP's image and description embeddings as the predictor for description quality, as formulated in (\ref{eq:clipscore}). Denoting $\frac{x}{|x|}$ as $\overline{x}$, we can express CLIPScore as
\begin{equation} \label{eq:clipscore}
   \text{max}\left({\overline{\textnormal{image}}} \cdot {\overline{\textnormal{description}}}, 0\right)
\end{equation}

SPURTS is different from CLIPScore since it only considers the description itself, without taking image information into account. The main goal of SPURTS is to detect fluency and style, and it can be written as 
\begin{equation} \label{eq:spurts}
\textnormal{median}_{\textnormal{layer}} \textnormal{max}_{\textnormal{head}} \  I_{\textnormal{flow}}(y_{w/o}, \theta),
\end{equation}
where $I_{\textnormal{flow}}$, which \citet{feinglass2021smurf} refer to as information flow, is normalized mutual information as defined in \citealt{witten2005practical}. For an input text without stop words, $y_{w/o}$, and a Transformer with parameters $\theta$ \citep{Vaswani-etal:2017}, SPURTS computes the information flow for each Transformer head at each layer, and then returns the layer-wise median of the head-wise maxima.

\subsection{Compatibility}

We first inspect the extent to which current referenceless metrics can capture whether a description is true for an image. SPURTS provides scores independent of the image and therefore inherently can't capture any notion of truthfulness.
In contrast, CLIPScore is trained to distinguish between fitting and non-fitting image--text pairs, returning a compatibility score. We test whether this generalizes to our experimental data by providing CLIPScore with the true descriptions written for each image and a shuffled variant where images and descriptions were randomly paired. As \figref{fig:clipscore-truthfulness}A demonstrates, CLIPScore rates the ordered pairs significantly higher compared to the shuffled counterparts ($\beta = 2.02$, $\text{SE} = .14$, $p<.001$),\footnote{Result from a linear effects analyses where the shuffled condition is coded as 0, and the ordered condition as 1.} suggesting that it captures image--text compatibility.

\subsection{Description Length Correlation}

Since the length of the description can already account for some of the variance of the BLV ratings, we further investigate whether description length is a general predictor for CLIPScore and SPURTS (see \figref{fig:clipscore-truthfulness}B). For CLIPScore, description length doesn't correlate with predicted quality of the description. This is likely a consequence of the contrastive learning objective, which only optimizes for compatibility but not quality. SPURTS scores, in contrast, significantly correlate with description length, which is aligned with the BLV ratings.

\subsection{Context Sensitivity}

Crucially, the descriptions were written and evaluated within contexts, i.e., their respective Wikipedia article, and previous work suggests that the availability of context should affect what constitutes a good and useful description.
Since current referenceless metrics can't integrate context, we expect that they shouldn't be able to capture the variation in the human description evaluations, and this is indeed what we find.

To investigate this hypothesis, we correlated sighted and BLV description evaluations with the CLIPScore and SPURTS ratings. As shown in \figref{fig:clipscore-blvsighted}, CLIPScore fails to capture any variation observed in the human judgments across questions. This suggests that, while CLIPScore can add a perspective on the compatibility of a text for an image, it can't get beyond that as an indication of how useful a description is if it's true for the image.

Like CLIPScore, SPURTS scores don't correlate with the sighted participant judgments (\figref{fig:clipscore-blvsighted}, bottom). However, specifically with respect to the overall rating, SPURTS scores show a significant correlation with the BLV participant ratings. While this seems encouraging, further analysis revealed that this correlation is primarily driven by the fact that both BLV and SPURTS ratings correlate with description length. The explained variance of the BLV ratings from description length alone is $0.152$ and SPURTS score alone explains $0.08$ of the variance. In conjunction, however, they only explain $0.166$ of the variance, which means that most of the predictability of SPURTS is due to the length correlation. This is further supported by a mixed effects linear regression analysis in which we fail to find a significant effect of SPURTS ($\beta = .80$, $\text{SE} = .44$, $p > .05$) once we include length as a predictor ($\beta = .64$, $\text{SE} = .15$, $p < .001$).\footnote{We assume random intercepts by participant and description, and we rescaled description length into $[0, 1]$.}

\begin{table}[tp]
\centering

\small
\begin{tabular}{@{} l@{ \ }l@{ \ } rrrr@{}}\toprule
  &&   Overall & Imagin. & Relev. & Irrelev. \\ 
  \midrule 
\multirow{2}{*}{\emph{BLV}} &
CLIPScore & 0.075   & 0.104         & 0.086     & 0.090       \\
& +Context & 0.201   & 0.182         & 0.202     & 0.142       \\
\midrule
\emph{Sighted,} & 
CLIPScore & $-$0.013 & 0.064         & 0.000     & $-$0.166     \\
\emph{no img}  & 
+Context & 0.238   & 0.315         & 0.190     & $-$0.019     \\
\midrule
\emph{Sighted,} & 
CLIPScore & 0.139   &              & 0.106     & $-$0.079     \\
\emph{w img} & 
+Context & 0.331   &              & 0.240     & 0.052       \\ \bottomrule
\end{tabular}
\caption{Comparison of the human rating correlations with the original context-independent CLIPScore and the context-sensitive adaptation, using the same CLIP embeddings. Missing cells were not experimentally measured by design (\secref{sec:exp-eval-sighted}). Across questions and participant groups, correlations improve. The CLIPScore correlations are a replication of \figref{fig:clipscore-blvsighted}.}
\label{tab:clipscorewcontext}
\end{table}

A further indication that SPURTS isn't capturing essential variance in BLV judgments is apparent from the negative correlation in the \emph{Irrelevance} question ($R=-0.25$). This suggests that SPURTS scores tend to be higher for descriptions that are judged to contain too much irrelevant information and low when participants assess the level of information to be appropriate. In the BLV responses, \emph{Irrelevance} is positively correlated with the \emph{Overall} ratings ($R=0.33$), posing a clear qualitative mismatch to SPURTS. Since what is considered extra information is dependent on the context, this is a concrete case where the metric's lack of context integration results in undesired behavior.

Finally, SPURTS' complete lack of correlation with sighted participant judgments further suggests that SPURTS is insufficient for picking up the semantic components of the descriptions. This aligns with the original conception of the metric, where a reference-based metric (SPARCS) is used to estimate semantic quality. 

Overall, our results highlight that SPURTS captures the BLV participants' preferences for longer descriptions but falls short in capturing additional semantic preferences, and is inherently inadequate for judging the truthfulness of a description more generally. CLIPScore can't capture any of the variation in BLV or sighted participant ratings, uncovering clear limitations.

\section{The Potential for Integrating Context into CLIPScore}

Can referenceless metrics like CLIPScore be made context sensitive? To begin exploring this question, as a proof of concept, we amend (\ref{eq:clipscore}) as follows:
\begin{multline} 
\label{eq:clipwcontext}
    \overline{\textnormal{description}} \cdot \textnormal{context} \ + \\   
    \textnormal{description} \cdot \left(\overline{\textnormal{image}}  - \overline{\textnormal{context}}\right)
\end{multline}
Here, quality is a function of (a) the description's similarity to the context (first addend) and (b) whether the description captures the information that the image adds to the context (second addend). These two addends can be seen as capturing aspects of (ir)relevance and imaginability, respectively, though we anticipate many alternative ways to quantify these dimensions.

\tabref{tab:clipscorewcontext} reports correlations between this augmented version of CLIPScore and our sighted and BLV participant judgments. We find it encouraging that even this simple approach to incorporating context boosts correlations with human ratings for all the questions in our experiment.
For the \emph{Irrelevance} question, it even clearly captures the positive correlation with BLV ratings, which is negative for both CLIPScore and SPURTS, indicating a promising shift. We consider this an encouraging signal that large pretrained models such as CLIP might still constitute a resource for developing future referenceless metrics.

However, despite these promising signs, there are also reasons to believe that CLIP-based metrics have other restrictive limitations. Due to CLIP's training, images are cropped at the center region and texts need to be truncated at 77 tokens \cite{radford2021learning}. 
CLIP relies on embeddings learned for each absolute token position in the text window and each patch position in the image. These can therefore not be easily extended to avoid any context or image cropping that is currently limiting CLIPScore.
Specifically for the purpose of accessibility, the information this removes can be crucial for determining whether a description is useful or not. For instance, our experiments show that the length of a description is an important indicator for description quality -- information lost in CLIP-based metrics. Moreover, this disproportionately affects the ability to encode the context paragraphs, which are often longer than a typical description. These decisions are therefore likely reflected in any resulting metric and should therefore be reconsidered when devising a new metric.

\section{Conclusion}

The context an image appears in shapes the way high-quality accessibility descriptions are written. In this work, we reported on experiments in which we explicitly varied the contexts images were presented in and investigated the effects of this contextual evaluation on participant ratings. These experiments reveal strong contextual effects for sighted and BLV participants. We showed that this poses a serious obstacle for current referenceless metrics, but we also see promise for future efforts since the inclusion of context to a prominent metric such as CLIPScore begins to address the disconnect from BLV needs.

\section*{Limitations and Ethics}

Our investigation focuses on whether a description is judged as fulfilling the purpose of accessibility and is therefore entirely based on utility considerations. However, generated image-based texts can also differ in  stylistic qualities such as grammaticality. \citet{kasai2021transparent} suggest a human annotation scheme that focuses on such dimensions, which together with our work provides a broad assessment of description quality.

To investigate the effect of context, we chose an experimental design where the same image can be placed in a variety of contexts and vice versa. Consequently, the images were complementary to the text, but the text could easily be understood without the image as well. Web accessibility guides suggest that only important images should receive alt descriptions and purely decorative images shouldn't receive any. Our image--context pairs are in between these extremes, and future work should explore how these effects vary depending on image--context relations. Similarly, we looked only at Wikipedia articles for providing context, but previous work has argued for paying attention to intricate differences between domains. While we expect the observed context effects to carry over to other domains such as social media, this is a matter for future investigation.

As our primary human evaluation method, we used 5-point Likert scales where a higher rating corresponded to a higher quality description across dimensions. Recently, \citet{ethayarajh2022human} argued against using Likert scales for comparing the performance of two systems on natural language generation tasks. Likert scales indeed come with challenges since the interpretation of the intervals is likely variable and asymmetric, posing a challenge for analysis (\citeauthor{jamieson2004likert} \citeyear{jamieson2004likert}, but see \citeauthor{carifio2008resolving} \citeyear{carifio2008resolving}, \citeauthor{norman2010likert} \citeyear{norman2010likert}). However, Likert scales are better supported in BLV optimized interfaces such as Google Forms, and were therefore chosen to allow a direct comparison between BLV and sighted participant judgments. To minimize potential artifacts due to the scale, we obtained multiple ratings from each participant and included by-participant random effects in the statistical analyses. Though not without challenges, Likert scales still provide the best method for quality assessments in accessibility-oriented comparisons.

All of our human subject experiments were conducted under IRB protocols. Most sighted participants spent between 20 and 30 minutes on the study and were paid \$6.15 (\$12.30--18.45/hr) over Amazon's Mechanical Turk. Most BLV participants completed the experiment between 1.5 and 2.5 hours (based on self reporting) and were paid \$75 in Amazon gift cards (\$30--50/hr, other gift cards being available upon request). The BLV study was thoroughly tested for its accessibility before it was distributed. All data were completely anonymized before analysis.

\section*{Acknowledgements}

This work is supported in part by a grant from Google through the Stanford Institute for Human-Centered AI and by the NSF under project REU-1950223. We thank our experiment participants for their invaluable input. We further thank Gabriel Poesia, Rachit Dubey, and Zhengxuan Wu for helpful discussions and comments.

\bibliography{Captioning_RM}
\bibliographystyle{acl_natbib}

\newpage

\appendix

\section*{Appendix}

\section{Guide to Supplementary Materials}

All data and code needed to replicate our results, along with additional analyses on the human subject data, are made available.\footnote{\url{https://github.com/elisakreiss/contextual-description-evaluation}} Specifically, all studies can be completed the exact way they appeared to participants, and our data and code enable replications of our graphs and statistical analyses, and provide further insights into the participant distributions, comments, and other additional analyses. Our materials also contain the code for the presented referenceless metrics adapted to our specific data. The README provides further details and points to the folders and files necessary to replicate our results.

\section{Implications for Other Referenceless Metrics}\label{app:others}

In the previous experiments, we established that the referenceless metrics CLIPScore and SPURTS can't get traction on what makes a good description when the images and descriptions are contextualized.
Other referenceless metrics such as UMIC \cite{lee2021umic} and QACE \cite{lee2021qace} face the same fundamental issue as CLIPScore and SPURTS due to their contextless nature. Like CLIPScore, UMIC is based on an image--text model (UNITER; \citealt{chen2020uniter}) trained under a contrastive learning objective. Similarly, it produces an image--text compatibility score solely by receiving a decontextualized image and text as input. 
QACE uses the candidate description to derive potential questions that should be answerable based on the image. The evaluation is therefore whether the description mentions aspects that are true of the image and not about which aspects of the image are relevant to describe. This again only provides insights into image--text compatibility but not contextual relevance.
Unfortunately, we are unable to provide quantitative results for these referenceless metrics since the authors haven't provided the code necessary (QACE), or the code relies on image features that can't be created for novel datasets with currently available hardware (UMIC, QACE).

In summary, the current context-independence of all existing referenceless metrics is a major limitation for their usefulness. This is a challenge that needs to be addressed to make these metrics a useful tool for advancing image-based NLG systems.

\section{Referenceless Metrics for Image-Based NLG Beyond Accessibility}

While we have specifically focused on the usefulness of referenceless metrics for image accessibility, this isn't the only potential purpose an image-based text might address. \citet{kreiss2022concadia} distinguish \emph{descriptions}, i.e., image-based texts that are written to replace the image, and \emph{captions}, i.e., texts that are intended to appear alongside images, such as tweets or newspaper captions. This suggests that the same text can be very useful for contextualizing an image but fail at providing image accessibility, and vice versa. 
To investigate this distinction, they asked participants to rate \texttt{alt} descriptions as well as image captions from Wikipedia according to (1) how much the text helped them imagine the image, and (2) how much they learned from the text that they couldn't have learned from the image. Descriptions were rated more useful for imagining the image, whereas captions were rated more useful for learning additional information. Captions used for contextualizing an image might therefore be another potential use domain for a referenceless metric such as CLIPScore. 

\begin{figure}
  \includegraphics[width=.48\textwidth]{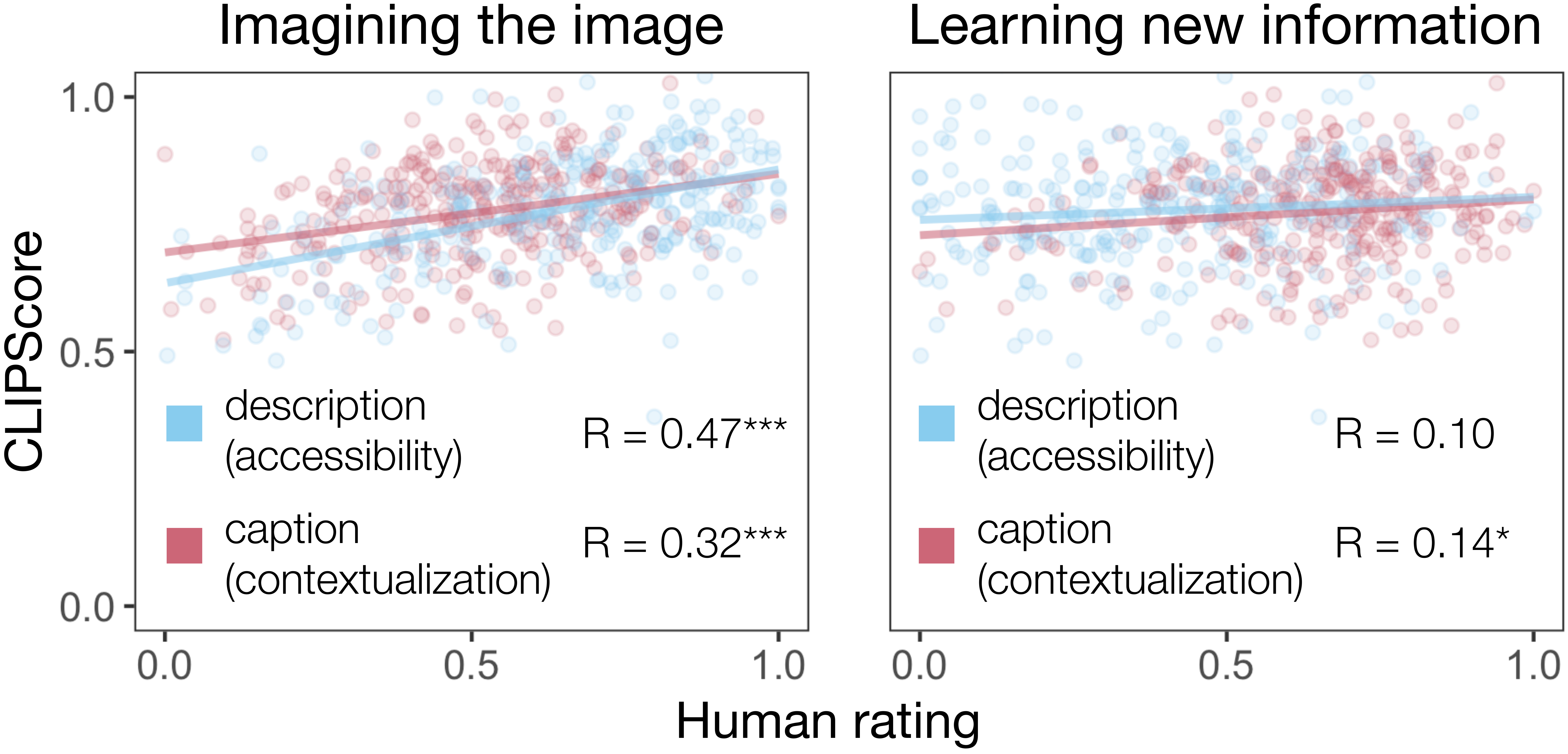}
  \caption{CLIPScore provides higher ratings for image-based texts that capture image content well. Whether the descriptions and captions provide additional information to the image content doesn’t affect the ratings.} 
  \label{fig:corr-clipscore-purpose}
\end{figure}

To see whether CLIPScore might be a promising resource for evaluating captions, we obtained CLIPScore ratings for the descriptions and captions in \citet{kreiss2022concadia}. CLIPScore ratings correlate with the reconstruction as opposed to the contextualization goal (see \figref{fig:corr-clipscore-purpose}), suggesting that CLIPScore is inherently less appropriate to be used for assessing caption datasets. This aligns with the original observation in \citet{hessel2021clipscore} that CLIPScore performs less well on the news caption dataset GoodNews \cite{biten2019gooda} compared to MSCOCO \cite{hessel2021clipscore}, a contextless description dataset.

Taken together, this is further evidence that the ``one-size-fits-all'' approach to referenceless image-based text evaluation is not sufficient for adequately assessing text quality for the contextualization or the accessibility domain.

\end{document}